\documentclass[final,3p,12pt]{elsarticle}

\usepackage{amssymb}
\usepackage{amsmath}
\usepackage{booktabs}
\usepackage{xcolor}
\usepackage{hyperref}
\usepackage{xurl}

\journal{Pattern Recognition}

\begin{document}

\begin{frontmatter}



\title{Multi-Stage Prototype Learning for Interpretable Time Series Classification}


\author[label1]{Bhavesh Kalisetti}
\author[label1]{Vincent Wang}
\author[label1]{Gaurav Ghosal}
\author[label1]{Maryam Bijanzadeh}
\author[label1,label2,label3,label4]{Reza Abbasi-Asl}

\affiliation[label1]{organization={Department of Neurology, University of California, San Francisco},
            addressline={}, 
            city={San Francisco},
            postcode={}, 
            state={CA},
            country={USA}}

\affiliation[label2]{organization={Department of Bioengineering and Therapeutic Sciences, University of California, San Francisco},
            addressline={}, 
            city={San Francisco},
            postcode={}, 
            state={CA},
            country={USA}}

\affiliation[label3]{organization={UCSF Weill Institute for Neurosciences},
            addressline={}, 
            city={San Francisco},
            postcode={}, 
            state={CA},
            country={USA}}   

\affiliation[label4]{Email: Reza.AbbasiAsl@ucsf.edu.}

\begin{abstract}
Deep learning methods are powerful tools in classifying multivariate time series data. Despite their high performance, these methods are hard to interpret, which diminishes their applications in high-risk domains such as healthcare. In this paper, we propose a novel multi-stage prototype learning framework for multivariate time series classification. By design, our framework identifies predictive temporal patterns in individual variables as well as cross-variable patterns that are highly predictive of each class. We validate our model on one simulated and four real-world datasets and demonstrate comparable accuracy to state-of-the-art methods while providing substantially improved interpretability through explicit, hierarchical prototype-based explanations. These explanations reveal both single-variable temporal patterns as well as cross-variable interactions that are most predictive for each class, providing insights into underlying mechanisms of the predictive model.

\end{abstract}







\end{frontmatter}



\section{Introduction}

Sensors embedded in wearable devices, industrial machinery, and medical monitoring 
systems continuously generate recordings of multiple variables over time. The task of 
assigning a class label to such multivariate time series and determining most predictive patterns for each class has become a central problem in pattern recognition. 
Applications of this kind span a broad range of domains, from financial analytics 
\cite{kwoncryptocurrency} and climate prediction \cite{climate} to disease detection 
\cite{deng2024interpretable} and human activity recognition. In many of these settings, 
and especially in healthcare \cite{li2018Healthcare}, a predictive model is only useful 
in practice if its decisions can be understood and verified. A practitioner who cannot 
inspect why a classifier reached a conclusion is unlikely to act on it. This places 
interpretability alongside accuracy as a primary design requirement, not an optional 
refinement \cite{murdoch2019definitions, rudin2019stop}.

Current approaches to multivariate time series classification broadly divide into two 
families. Feature-based methods, including shapelet discovery 
\cite{ye2009shapelet, grabocka2015shapelet} and bag-of-patterns representations 
\cite{weaselmuse, baydogan2015symbolic}, extract human-readable subsections 
of signals that are maximally representative of each class. This allows a practitioner to 
visually verify what the model has learned. They have historically been the preferred 
choice when interpretability is required. However, their computational cost grows steeply 
with sequence length and the number of variables, limiting their applicability to the 
large-scale recordings in modern biosensing measurements. Deep 
learning methods handle such data more efficiently and achieve high classification 
accuracy \cite{ismail2019deep}. For example, recurrent, convolutional, transformer-based, and hybrid architectures 
have all demonstrated strong performance on standard multivariate benchmarks 
\cite{ruiz2021great, zerveas2021transformer}. The fundamental difficulty is that their predictions are often hard to interpret for domain expert. Post-hoc 
attribution methods, including gradient-based saliency maps \cite{qi2020visualizing, 
samek2021explaining} and Shapley value decompositions \cite{chen2023algorithms}, can 
highlight which input regions influenced a given output. However, these explanations are 
constructed after training and are not guaranteed to faithfully reflect the model's true 
decision logic \cite{rudin2019stop}. This becomes an issue particularly for safety-critical applications such as clinical monitoring.

Prototype-based learning offers a principled solution  \cite{li2018prototypes}. Rather than approximating 
explanations from a fixed black-box model, it builds interpretability directly into the 
classification architecture: a small set of representative patterns, or prototypes, is 
learned during training, and every prediction is made by explicit comparison to these 
exemplars. This form of case-based reasoning was first developed for image recognition 
\cite{chen2019looks, li2018prototypes} and subsequently extended to time series 
by Gee et al.\ \cite{gee2019prototypes} and Ming et al.\ \cite{ming2019interpretable}. Despite their appeal, 
existing prototype methods for multivariate time series share two structural limitations. 
First, they encode all variables jointly in a single shared latent space. This design 
makes it difficult to attribute a prediction to a specific variable or to identify 
patterns that arise from the interaction of multiple variables. Second, 
they treat the number of prototypes as a hyperparameter to be tuned by trial and error, 
with no principled basis for selection.

\begin{figure*}[h]
    \centering
    \includegraphics[width=\textwidth]{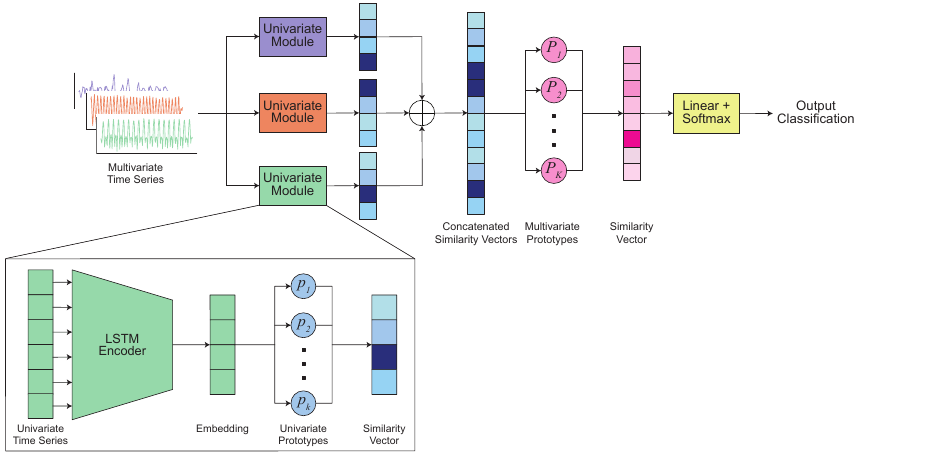}
    \caption{MMPL Architecture. The multivariate time series is split into univariate modules, each of which contain an LSTM encoder to learn lower-dimensional representations, as well as a univariate prototype layer to capture variable-level patterns. Similarity vectors outputted from each univariate module are concatenated to produce the input to the multivariate module. Another prototype layer learns class-level representations, and classifications are produced through softmax.}
    \label{fig:model}
\end{figure*}

Here, we address both limitations with a modular, multi-stage prototype learning framework 
(MMPL) (Figure~\ref{fig:model}). In the 
first stage of MMPL, each variable of the multivariate time series is encoded independently by a 
dedicated LSTM, and a separate prototype layer learns a compact set of temporal patterns 
for that variable alone. The number of prototypes per variable is selected automatically 
using silhouette analysis \cite{rousseeuw1987silhouette} applied to the pre-trained 
latent space. In the second 
stage, the per-variable similarity scores are concatenated and matched against a second 
prototype layer, one prototype per class. These prototypes captures how variable-level patterns 
co-occur to define each class label. This two-stage design supports interpretation at two 
levels of granularity: which patterns appear within each individual sensor channel, and 
how those patterns combine across channels to produce a classification decision.
To ensure that the per-variable latent spaces are well-structured for 
prototype learning, each variable encoder is pre-trained using contrastive learning 
\cite{hadsell2006contrastive}. This produces tight and well-separated clusters in each variable's latent space. As our 
ablation study demonstrates, this substantially reduces prototype redundancy and improves 
both accuracy and explanation clarity relative to joint end-to-end training. 

We validate MMPL on one synthetic dataset with known ground-truth cross-variable 
structure and four real-world datasets: three benchmarks from the UEA Multivariate Time 
Series Classification Archive \cite{bagnall2018uea} and the WESAD multi-modal biosensing 
dataset \cite{wesad}. Across all five settings, MMPL achieves accuracy 
comparable to the best single-model approaches, including WEASEL+MUSE 
\cite{weaselmuse} and MLSTM-FCN \cite{mlstmfcn}, while 
producing prototype representations that align with domain knowledge.

\section{Related Works}
Multivariate time series classification is a widely studied problem, with methods proposed in a variety of domains.
In this section, we provide an overview of prior works with an emphasis on interpretability. 
We categorize existing methods into feature-based approaches and deep learning approaches.

\subsection{Feature-based Approaches}
Feature-based approaches aim to extract features from the time series to perform classification.
One important class of feature-based approaches are shapelet methods.
A shapelet is a subsection of a time series that is "maximally representative" of a class \cite{ye2009shapelet}.
Since they extract relevant features of a time series, shapelet methods are typically interpretable, indicating their usefulness for healthcare applications.
Indeed in \cite{ghalwash2012shapelet}, shapelets are used for early and patient-specific classification of multivariate time series (MTS). Grabocka et.al.
 \cite{grabocka2015shapelet} demonstrates an efficient method for discovering shapelets by using an online clustering pruning technique. 
Another prevalent class of feature-based approaches is bag-of-patterns methods, which utilize sliding windows to generate representations of subsequences of a time series and then create a histogram of the symbols. Schäfer and Leser \cite{weaselmuse} creates a representation for an MTS by applying a sliding window to each variable and filtering out non-discriminative features.
Meanwhile,  Gokce et.al. \cite{baydogan2015symbolic} considered all variables of the MTS simultaneously when generating a representation using a tree-based ensemble.
Despite their usefulness in providing interpretability, feature-based approaches can be computationally expensive, as the set of subsequences and sliding windows grows exponentially when the sequence length of the time series increases. A comprehensive benchmark of multivariate time series classification methods across the UEA archive \cite{ruiz2021great} confirms that while feature-based methods achieve competitive accuracy on structured datasets, their computational cost limits applicability to the long, high-dimensional time series common in modern biosensing applications.

\subsection{Neural Network-Based Approaches}
Despite achieving high accuracy in classification, neural network-based approaches often lack interpretability due to their inherent black-box nature. Post-hoc explanation methods such as saliency maps, gradient-weighted class activation mapping \cite{qi2020visualizing, samek2021explaining}, and Shapley value-based attribution \cite{chen2023algorithms} can highlight which input regions influenced a prediction, but these explanations are generated after training and are not guaranteed to faithfully reflect the model's true decision logic. For high-stakes clinical applications, intrinsically interpretable models, those whose decision process is transparent by design rather than approximated after the fact, are preferable \cite{rudin2019stop}. As such, some prior works aim to combine neural networks with feature extraction methods.
For example, \cite{tang2020fewshot} uses neural networks to learn representative and discriminative shapelets.

Other studies on prototype learning for image classification \cite{chen2019looks, li2018prototypes, ao2022cross} or visual pattern mining \cite{shi2026rethinking, zhang2022zero} have focused on building networks that learn a set of prototype image patches and classify or mine inputs according to their similarity to these prototypes. This provides a form of case-based reasoning that is directly interpretable: a prediction is explained by identifying the training examples that most resemble the input. Gee et al. \cite{gee2019prototypes} adapted prototype learning to time series data, learning lower-dimensional representations of univariate and multivariate time series followed by a prototype layer in the latent space, and demonstrated its utility for clinical time series classification. Ming et al. \cite{ming2019interpretable} further extended prototype-based sequence learning with a steering mechanism that allows domain experts to interactively refine prototypes toward more meaningful representations. Both \cite{gee2019prototypes} and \cite{ming2019interpretable} employ three regularization terms, prototype similarity, prototype diversity, and encoded space coverage.
The prototype similarity term ensures that the prototypes are similar to training point embeddings.
The diversity term encourages prototypes to be far from each other.
Lastly, the coverage term forces prototypes to represent the entirety of the learned latent space.
Our method builds upon these works with two levels of prototype learning, allowing for the identification of cross-variable patterns as well as individual feature relevance.

More recently, transformer-based architectures have achieved state-of-the-art accuracy on multivariate time series classification benchmarks by capturing long-range temporal dependencies through self-attention mechanisms \cite{zerveas2021transformer}. While these models deliver strong predictive performance, their attention weights do not provide the kind of explicit, human-verifiable pattern representations that prototype-based methods offer. Nauta et al. \cite{nauta2021neural} demonstrated that hierarchical prototype trees can scale interpretable prototype learning to fine-grained recognition tasks while preserving classification accuracy, motivating our two-stage hierarchical design. A comprehensive review of deep learning approaches for time series classification is provided by Fawaz et al. \cite{ismail2019deep}, which we use as a reference for situating MMPL within the broader methodological landscape.

While the works described above demonstrate the viability of prototype-based interpretability for time series, they share a fundamental limitation: a single shared latent space for all variables cannot distinguish which variable drives a prediction, nor can it identify the cross-variable interaction patterns that are central to multi-modal biosensing. Our framework addresses both limitations through hierarchical prototype learning with per-variable LSTM encoders, contrastive pre-training for structured latent spaces, and a principled silhouette-based method for prototype count selection that removes a key hyperparameter from the practitioner's hands.

\section{Methodology}

\subsection{Model Architecture}

At a high level, MMPL operates in two stages. In the first stage, each variable in the multivariate time series is independently encoded by a dedicated LSTM and associated with a small set of univariate prototypes, each representing a recurring temporal pattern within that variable. The similarity of an input to each prototype is computed and collected into a variable-level similarity vector. In the second stage, the similarity vectors from all variables are concatenated and matched against a set of multivariate prototypes, one per class, that capture how variable-level patterns co-occur to define each class label. This hierarchical design enables interpretation at two levels of granularity: which patterns appear in individual variables, and how those patterns combine across variables to produce a classification.

Let $D = \{((\mathbf{x}_i^{(t)})_{t=1}^{T}, y_i)\}_{i=1}^{n}$ be the training dataset with sequence length $T$, $\mathbf{x}_i^{(t)}\in\mathbb{R}^{d}$, and class labels $y_i\in\{1,...,K\}$ for each $i\in\{1,...,n\}$.
Our model architecture is a two-stage encoder-prototype network.
The first stage consists of $d$ univariate modules, each of which contains an LSTM encoder $f: \mathbb{R}^T\to\mathbb{R}^m$ and a prototype network $p: \mathbb{R}^m\to\mathbb{R}^{k}$.
These modules separate the $d$-dimensional multivariate time series input into a set of $d$ univariate time series.
Each univariate module performs dimensionality reduction using its encoder, learning an embedding space and extracting relevant features.
The embeddings are then passed into a prototype network, where the module learns $k$  \textit{univariate prototype vectors} $\mathbf{p}_1,...,\mathbf{p}_k\in\mathbb{R}^m$.
Notably, we allow the number of prototypes in each univariate module to vary.
The prototype network $p$ computes a similarity vector $\mathbf{s}\in\mathbb{R}^k$ using a similarity function applied to the encoded input and all $k$ prototypes.
In this method, the similarity between two vectors $\mathbf{a}$ and $\mathbf{b}$ is computed to be $\frac{1}{||\mathbf{a} - \mathbf{b}||_{2}^{2} + \epsilon}$, where $\epsilon$ is for numeric stability.

We construct the input to the second stage of our model by concatenating the $d$ similarity vectors obtained from the univariate modules.
The resulting concatenated vector $\textbf{c} = \textbf{s}_1\oplus...\oplus \textbf{s}_d$ is of length equal to the sum of the number of prototypes in each univariate module.
This vector is passed into the final prototype network $P: \mathbb{R}^{\sum_{i=1}^{d}k^i}\to \mathbb{R}^K$ , which learns $K$ \textit{multivariate prototype vectors} and produces a similarity vector $\mathbf{S}\in\mathbb{R}^{K}$ using the same similarity function as the univariate modules.
The similarity vector is passed into a fully-connected layer $w: \mathbb{R}^K\to\mathbb{R}^K$ followed by a softmax layer $s$ to produce a probability distribution over $K$ classes.

The primary advantage of our model over prior work is the pattern-level interpretability in the first stage and the composition of the relevant patterns into the multivariate interactions in the second stage.
In this way, the two-stage prototype network's predictions are transparent to interpretation at all layers.
Unlike previous approaches that rely on a single latent space for the entire multivariate time series, this method learns distinct latent spaces and prototypes for each variable. This key difference allows for the identification of unique variable-specific patterns that would otherwise remain hidden.
Moreover, the separation of variables in our framework, which is achieved by computing silhouette scores for different variations of $k$-means clustering,allows us to parameterize the number of univariate prototypes in each module.
This approach allows us to interpret univariate prototypes as distinct patterns within each variable. The multivariate prototypes then reveal how these individual variable patterns combine and interact to produce a specific classification.
Second, our model allows for greater flexibility when choosing hyperparameters and training.
Because each univariate module is learned independently, different variables can have different latent space sizes, number of univariate prototypes, and even encoder architectures.
This is highly relevant to the aforementioned multi-modal biosensing problem, where each biosensor likely has different characteristics.
Third, while prior work varies the number of prototypes as a hyperparameter, our model learns $K$ multivariate prototypes and parameterizes the number of univariate prototypes in each variable by computing silhouette scores.
This increases interpretability by allowing the model to learn a representation for each class label.

\subsection{Objective Function/Training Procedure}
\subsubsection{Encoder Pre-training via Contrastive Learning} 
In previous encoder-prototype works for multivariate time series classification \cite{gee2019prototypes}, the encoder and prototype network are trained jointly, resulting in a cost function that incorporates classification loss and reconstruction loss along with various regularization terms.
As an improvement, we propose a separate pre-training step for the univariate encoders using contrastive learning \cite{hadsell2006contrastive}.

In this pre-training step, pairs of training examples $(\mathbf{x}_1, \mathbf{x}_2)$ are first labeled according to a similarity criterion: pairs sharing the same class label are marked "similar" ($Y=0$), while pairs from different classes are marked "dissimilar" ($Y=1$). Each pair is passed through the encoder network $f$, which computes their respective embeddings ($(f(\mathbf{x}_1), f(\mathbf{x}_2))$). The contrastive loss [HADSELL2006, CHEN2020] is then minimized:
\begin{equation} \label{eq:contrastive-loss}
    L_{c} = \frac{1}{2}(1-Y)(D_{W})^{2} + \frac{1}{2}(Y)\{\max{(0, m - D_{W})}\}^2
\end{equation}
where $D_W = ||f(\mathbf{x}_1 - f(\mathbf{x}_2)||_{2}$ and $m>0$ is a margin hyperparameter controlling the minimum separation between dissimilar pairs. By structuring training in this way, we ensure that examples from the same class are mapped to nearby regions of the latent space while examples from different classes are separated by at least margin $m$, producing well-defined clusters that represent distinct temporal patterns and substantially simplifying the subsequent prototype learning stage.

\subsubsection{Univariate Prototypes Training}
Once the encoders have been pre-trained, the univariate prototype networks $p_1, ..., p_d$ are trained.
The univariate prototypes are learned through classification.
We concatenate the $d$ similarity vectors produced by the univariate modules and pass the resulting vector through a fully-connected layer $g$ and a softmax layer to determine a classification output.
We train this stage of the model minimizing cross entropy loss.
Furthermore, we also leverage three regularization terms to improve the interpretability of the learned univariate prototypes.
We adopt two of the regularization terms from \cite{li2018prototypes}:

$$R_1 = \frac{1}{k}\sum_{j=1}^{k}\min_{i\in[1,n]}||\mathbf{p}_j-f(\mathbf{x}_i)||_{2}^{2}$$
$$R_2 = \frac{1}{n}\sum_{i=1}^{n}\min_{j\in[1,k]}||\mathbf{p}_j-f(\mathbf{x}_i)||_{2}^{2}$$

$R_1$ denotes the prototype similarity regularization term, which penalizes the distance between each prototype and its nearest training example.
This term ensures that learned prototypes are similar to at least one training example, allowing us to use the nearest training example as a projection of the prototype back to the input space.
$R_2$ denotes the encoded space coverage regularization term, which penalizes the distance between each training example and its nearest prototype.
This term ensures that learned prototypes cover the entire latent space, which is essential as the interpretability and performance of prototype learning depends on all input points being well-represented.

We also include the prototype diversity regularization term from \cite{ming2019interpretable}:
$$R_3 = \sum_{i=1}^{k}\sum_{j=i+1}^{k}\max{(0, d_{min}-||\mathbf{p}_i-\mathbf{p}_j||_2)^2}$$
where $d_{min}$ is a tunable hyperparameter that functions as a margin.
This term penalizes prototypes that are similar, helping reduce redundancy in learned prototypes.
The three regularization terms are summed up across the univariate modules and the resulting loss function for the first stage is shown in Equation \ref{eq:univariate-prototypes-loss}:

\begin{equation} \label{eq:univariate-prototypes-loss}
    \begin{aligned}
        & L(p_1, ..., p_d, g), X) = CE(p_1, ..., p_d, g, X)\\
        & + \sum_{i=1}^{d} R_1(p_i, X) + R_2(p_i, X) + R_3(p_i)
    \end{aligned}
\end{equation}

\subsubsection{Choosing the Number of Univariate Prototypes}

Previous works parameterize the number of prototypes by increasing it until performance gains are minimal.
Although this can be effective, we provide a more robust process to select the number of prototypes that is grounded in the structure of the data in the latent space by using silhouette analysis \cite{rousseeuw1987silhouette}.
Typically used in clustering, a silhouette value is a measure of how similar a point is to its own cluster and how dissimilar it is to other clusters.
Our framework leverages the latent embeddings learned by the pre-trained encoders for silhouette analysis.
These clusters are also generally tight thanks to the nature of contrastive learning, ensuring that the training data can be divided into clusters properly.
For each variable, we apply k-means clustering \cite{lloyd1982kmeans}, experimenting with different numbers of clusters (k).
We select the largest value of $k$ for each variable that does not result in a significant decrease in silhouette score.
As a result, we choose the number of univariate prototypes that results in maximum expressiveness while maintaining an optimal separation of the training data.
Evaluating our model on various datasets, we find the $k$-value selected by this method such that it does not sacrifice accuracy.

An additional optimization we use as a result of pre-training encoders is the $k$-means++ algorithm \cite{arthur2007kmeans} to initialize univariate prototypes.
Because the latent spaces are fixed before the prototypes are learned, we initialize the prototypes to maximize their distance to each other, as opposed to random initialization done by previous works.
This yields faster convergence for the first stage of our model.

\subsubsection{Multivariate Prototypes Training}

Once the univariate prototypes have been learned for each module, the multivariate prototype network $P$ is trained, freezing all other parameters in the model.
The cost function for this stage follows the same structure as the previous stage, using cross entropy loss alongside the three additional regularization terms, as shown in Equation \ref{eq:multivariate-prototypes-loss}

\begin{equation} \label{eq:multivariate-prototypes-loss}
    \begin{aligned}
        & L(P) = CE(P, X)\\
        & + R_1(P^{(1)},...,P^{(K)}, X)\\
        &+ R_2(P^{(1)},...,P^{(K)}, X)\\
        &+ R_3(P^{(1)},...,P^{(K)})
    \end{aligned}
\end{equation}

\section{Results}

In this section, we evaluate our model on five datasets: a simulated dataset that incorporates cross-variable patterns, and three time series datasets from the UEA Time Series Repository \cite{bagnall2018uea}. These datasets include two multivariate accelerometer data sets during different types of physical activity and a pen trajectory data set for character identification. Finally, we will demonstrate the utility of MMPL in investigating WESAD \cite{wesad}, a multi-modal biosensing dataset which consists of both physiological \& accelerometer data. In each dataset, we assess the capabilities of our model in terms of accuracy as well as interpretability and marker discovery.

We compare our model's classification accuracy on the simulated dataset and UEA Time Series Repository datasets to two strong baseline models: WEASEL+MUSE \cite{weaselmuse} and MLSTM-FCN \cite{mlstmfcn}, which are among the strongest individual feature-based and deep learning-based algorithms for multivariate time series classification \cite{ruiz2021great}. We note that the current state-of-the-art on the UEA archive is held by ensemble methods such as HIVE-COTE 2.0 \cite{middlehurst2021hive}, which achieves near-perfect accuracy on many benchmarks but provides no interpretability. Our goal is not to surpass such ensembles, but to demonstrate that prototype-based interpretability can be achieved without substantial sacrifice in accuracy relative to the best single-model approaches.
For WESAD, we compared our model's classification accuracy to the findings described in \cite{wesad}, which contains accuracies for multiple traditional machine learning algorithms.
Furthermore, for our simulated dataset, we performed ablation studies to demonstrate how contrastive learning and variable separation impact performance.

\subsection{Performance of the MMPL on Simulated Dataset}

\subsubsection{Dataset Formulation}

We constructed a simulated dataset with the multi-modal biosensing problem in mind.
We focused on three main aspects of the problem in our design.
First, some sensors may produce data that is irrelevant to the classification task.
The model should be able to ignore these irrelevant variables.
Second, cross-variable patterns may be present in data, requiring a model to be able to consider data from multiple sensors simultaneously in order to perform classification.
Third, data comes from a diverse set of sensors, each of which may have vastly different patterns.
We designed the simulated dataset to explicitly contain these three characteristics.

\begin{figure*}
    \centering
    \includegraphics[width=.8\textwidth]{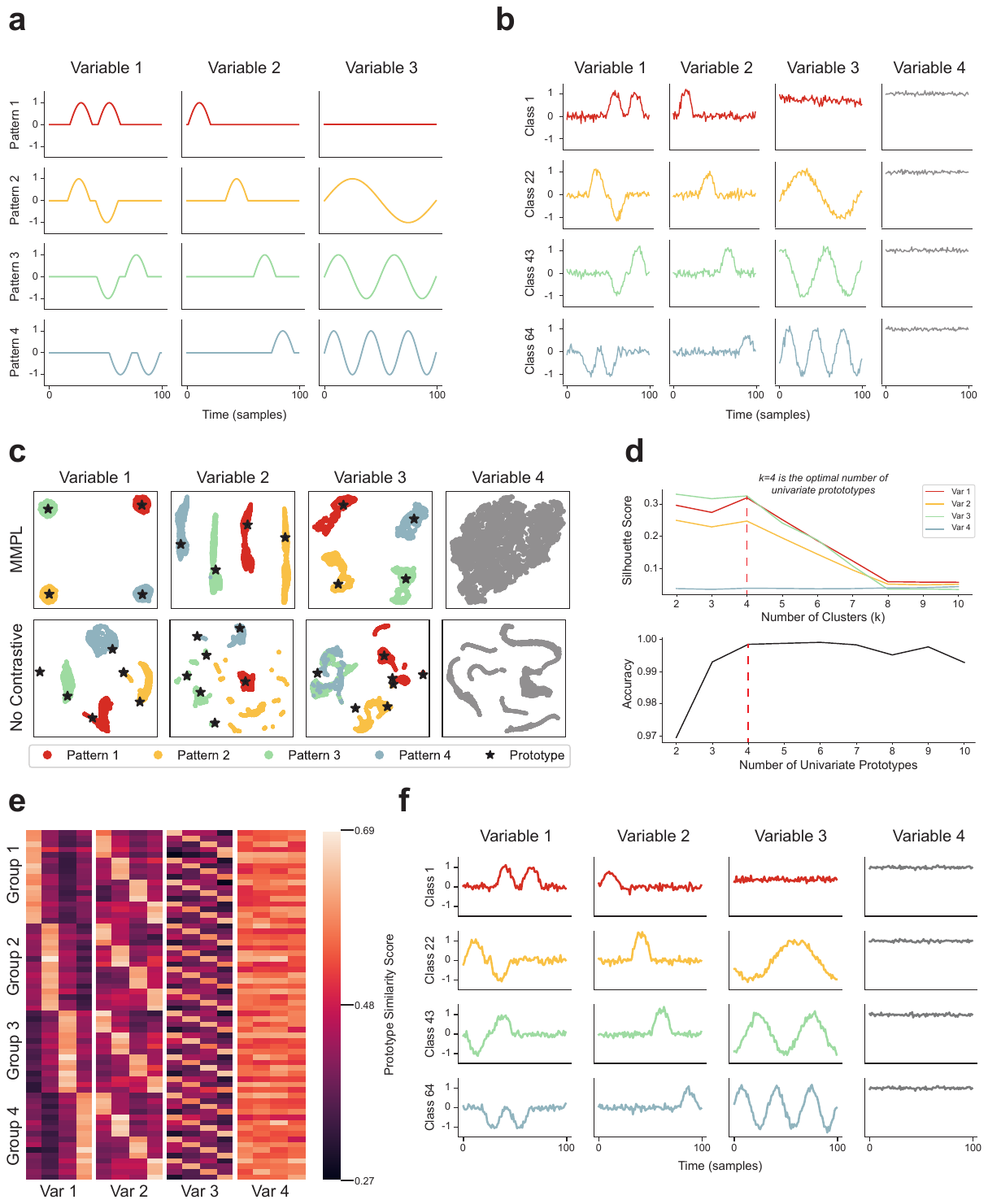}
    \caption{MMPL achieves pattern-level separation and learns class-level representations on the simulated dataset. \textbf{a}, The underlying patterns used to generate the simulated dataset. \textbf{b}, Training points generated by randomly sampling the underlying patterns. \textbf{c}, A comparison of the learned latent spaces with and without contrastive learning, visualized using UMAP. Contrastive learning yields clear separation of variable-level patterns and an efficient number of univariate prototypes. \textbf{d}, The number of univariate prototypes is determined by examining how silhouette score changes as the $k$ in $k$-means clustering varies. $k=4$ is the largest number before a significant drop in silhouette score, indicating that $4$ is the optimal number. \textbf{e}, A heatmap of the learned multivariate prototypes, which are unique and are made up of "one-hot encoded" vectors for each variable. \textbf{f}, Learned prototypes projected back to the input space, which are consistent with dataset design.}
    \label{fig:simulated_fig}
\end{figure*}

In the simulated dataset, the first three variables contain patterns relevant for classification, while the fourth variable contains random noise irrelevant to the task.
Each of the three relevant variables contains four different patterns.
The combination of variable-level patterns in a time series determines its class, resulting in $4^{3}=64$ different classes.
Therefore, all three relevant variables are necessary in order to correctly classify a data sample.
Each of the three relevant variables was simulated such that it represents a different time series pattern.
The first variable contains "shift-invariant" patterns, meaning the pattern is independent of its location within the time series.
The second variable contains "shift-variant" patterns, meaning the location of the pattern within the time series is relevant towards classification.
The third variable contains sinusoidal data, with representing different frequency patterns.
Figure~\ref{fig:simulated_fig}a shows the four patterns for each of the relevant variables and Figure~\ref{fig:simulated_fig}b shows sample points generated by adding noise to the combinations of variable-level patterns.

We generated samples by applying random noise to each pattern over 100 time steps.
The train and test set each contained 100 samples per class for a total of 6400 samples each.

\subsubsection{Classification Accuracy and Interpretability Evaluation}

Our interpretable model achieves a classification accuracy of 98.5\% on the test set, comparable to the performance reported in
MLSTM-FCN \cite{mlstmfcn} (99.8\% classification accuracy). 
We investigate the interpretability of our model by visualizing the latent space along with the learned univariate and multivariate prototypes.
Figure~\ref{fig:simulated_fig}c shows a 2 dimensional UMAP \cite{mcinnes2018umap} visualization of the latent space for each variable, along with the learned univariate prototypes depicted by black stars.
The first three variables exhibit four distinct clusters corresponding to the four distinct patterns in each variable.
The latent space of the fourth variable does not offer any distinct cluster, with the embeddings forming one large cluster.
The structure of the latent space demonstrates our model's ability to distinguish between the patterns in each variable. Notably, the fourth variable, which contains only random noise, forms a single undifferentiated cluster, confirming that our model does not impose spurious structure on uninformative inputs.

Figure~\ref{fig:simulated_fig}d shows a plot of silhouette score across different values of $k$.
The maximum value of $k$ without compromising on the separability of the clusters is 4, which aligns with the value determined from comparing classification accuracy to the number of univariate prototypes.
With $k$ = 4 for all variables, our model learns one univariate prototype per cluster in each of the three relevant variables.
The position of the univariate prototypes indicates our model's ability to learn representative examples of each variable-level pattern. The alignment between the silhouette-selected value of $k=4$ and the number of ground-truth patterns in each variable suggests that our prototype count selection procedure is a reliable, data-driven alternative to manual hyperparameter tuning.

As there are 64 classes in the simulated dataset, we set the number of multivariate prototypes to 64.
The learned multivariate prototypes are shown in Figure~\ref{fig:simulated_fig}e.
The model's multivariate prototype selects a single univariate prototype from each of the three relevant variables.
For the irrelevant fourth variable, the multivariate prototypes contain uniformly distributed values, indicating the model does not select any univariate prototype over the others.
Our model explicitly recovers the simulated dataset's design through its prototypes.
The learned multivariate prototypes are also distinct from one another, demonstrating the model's ability to learn a prototype for each class label.

Therefore, our model recovers the cross-variable patterns present in the data in an intuitive way. The uniform prototype similarity values assigned to the fourth variable across all multivariate prototypes confirm that MMPL successfully identifies and discards irrelevant sensor channels without any explicit supervision to do so.
In Figure~\ref{fig:simulated_fig}f, multivariate prototypes for select classes are visualized in the input space using nearest neighbor projection.
These projected prototypes are consistent with the design of the simulated dataset, demonstrating that the nearest-neighbor projection provides a faithful and interpretable mapping from the latent prototype back to the original time series domain.


\subsubsection{Ablation Studies}

To examine how variable separation and contrastive learning affect interpretability and performance, we perform two ablation studies.
First, we train a one-stage encoder-prototype network without contrastive learning on the simulated dataset.
This model consists of a singular LSTM encoder followed by a prototype network and a softmax layer to produce classifications, which matches the architecture of previous interpretable prototype learning methods \cite{gee2019prototypes}, \cite{ming2019interpretable}.
After training, we observe a test accuracy of 54.2\%, significantly lower than our model's accuracy.
Furthermore, the lack of variable separation makes the model less interpretable, as we cannot determine whether the model was able to learn the cross-variable patterns present in the data.
Additionally, interpreting the model's single latent space proves challenging, as it cannot adequately encode all relevant information from the input, given the nature of the simulated dataset. By contrast, separating variables into distinct latent spaces enables the model to capture these crucial cross-variable interactions effectively.

Second, to observe how contrastive learning improves accuracy and interpretability, we train a two-stage encoder-prototype network with variable separation but discard pre-training the encoders using contrastive learning. 
In each univariate module, we train the encoder and prototype network jointly using cross entropy loss and regularization terms similar to \cite{li2018prototypes} and \cite{gee2019prototypes}.
After training, we observe a test accuracy of 74.8\%, which is higher than the one-stage model but still lower than our two-stage model with contrastive learning.
Figure~\ref{fig:simulated_fig}c bottom row, shows the latent spaces and univariate prototypes learned by this altered model in comparison to MMPL.
The model fails to form four distinct clusters for each pattern.
In variable 2, the model splits pattern 2 and pattern 3 into several small clusters.
In variable 3, the model fails to separate pattern 3 and pattern 4.
A poorer quality latent space also requires more univariate prototypes to cover the encoded space, making training more difficult and increasing the chance of prototype redundancy.
This makes interpretations of the prototypes more ambiguous compared to when contrastive learning was used where the model was also able to learn one univariate prototype per cluster.
Downstream, the multivariate prototypes also become more difficult to interpret in terms of the combinations of patterns across variables forming each class.

Our findings demonstrate that pre-training with contrastive learning leads to significant improvements in performance, latent space quality, and model interpretability. The degraded clusters produced without contrastive pre-training require more univariate prototypes to cover the encoded space, increasing the risk of redundancy and making the learned multivariate prototypes harder to interpret.



Together, these ablation results isolate the contribution of each design choice: variable separation is necessary to handle heterogeneous multi-variable data, and contrastive pre-training is necessary to produce the well-structured latent spaces that make prototype learning meaningful. The broader implications of these findings for model design are discussed in Section~5.

\begin{table}[ht]
\centering
\caption{Accuracy Performance on UEA Archive Benchmark Datasets. Bold indicates top performance.}
\label{table:uea_accuracies}
\begin{tabular}{@{}lccc@{}}
\toprule
\textbf{Model} & \textbf{Character Trajectories} & \textbf{Epilepsy} & \textbf{BasicMotions} \\ \midrule
MLSTM-FCN & \textbf{99.3\%} & 96.4\% & \textbf{100.0\%} \\
WEASEL+MUSE & 99.0\% & \textbf{99.3\%} & \textbf{100.0\%} \\ \midrule
\textbf{MMPL (Ours)} & 96.6\% & \textit{96.9\%} & 99.8\% \\ \bottomrule
\end{tabular}
\end{table}

\subsection{Prototype learning during seizure mimicking behavior}

We next evaluated our model on tri-axial accelerometer data during four activities including a seizure mimicking behavior. We used the dataset from the UEA Time Series Repository \cite{epilepsy}, containing accelerometer measurements from the dominant wrist of subjects whilst completing four activities: seizure mimicking, running, walking, and sawing.
The sampling frequency was 16 Hz and each multivariate time series was truncated to the length of the shortest series, which was approximately 13 seconds.
The classification objective is to predict the activity from the accelerometer readings.

The predictive model achieves a classification accuracy of 96.9\% on the test set. Accuracy comparisons are shown in Table~\ref{table:uea_accuracies}. Notably, our model achieves comparable accuracy to MLSTM-FCN and slightly lower accuracy than WEASEL +MUSE. This result indicates that we do not necessarily need to sacrifice accuracy to achieve interpretability and that our interpretable model structure does not hinder performance.

\begin{figure*}
    \centering
    \includegraphics[width=\textwidth]{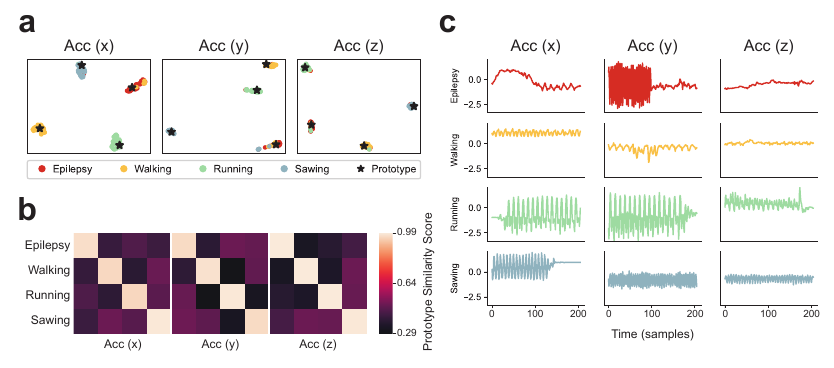}
    \caption{MMPL learns representative examples of all four classes of the Epilepsy dataset. \textbf{a}, UMAP visualization of the learned latent spaces shows the four classes are distinguishable in all variables. \textbf{b}, MMPL selects exactly one univariate prototype per variable when constructing multivariate prototypes. \textbf{c}, Projected prototypes match domain knowledge. The prototype for Walking contains signals with low amplitude, while the prototype for Running contains signals with higher amplitude and frequency.}
    \label{fig:epilepsy_fig}
\end{figure*}

We visualize the latent spaces for our model in Figure~\ref{fig:epilepsy_fig}a and observe distinct clusters for each class across all three accelerometer axes, indicating that the four activities produce separable signal patterns in every variable. This per-variable separability suggests that each accelerometer axis independently captures discriminative motion information, a finding that is directly revealed by MMPL's variable-level latent spaces but would be invisible in a single shared multivariate latent space.
Here, we select the number of univariate prototypes for each variable to be 4. We observe that each cluster in the latent space has a single univariate prototype assigned to it, validating our model's ability to learn a compact, non-redundant set of representative patterns at the variable level.

We set the number of multivariate prototypes to 4, as there are 4 classes in the dataset. Visualizations of the learned prototypes are presented in Figure~\ref{fig:epilepsy_fig}b.
Once again, the multivariate prototype vectors select one univariate prototype per variable for all 3 variables, implying the data are completely distinguishable in all variables.
Figure~\ref{fig:epilepsy_fig}c shows the prototypes projected back to the input space. The prototype for Walking contains signals with low amplitude, the prototype for Running contains signals with higher amplitude and frequency, and the prototype for seizure mimicking activity contains stochastic and non-stationary patterns. These signal characteristics are consistent with known biomechanical differences between these activities. This suggests that MMPL's learned prototypes are not arbitrary latent constructs but correspond to physically meaningful motion signatures that a domain expert can directly verify.

These results demonstrate that MMPL's prototypes align with known biomechanical signatures of the four activities, validating the framework's ability to recover domain-meaningful patterns without manual feature engineering. The broader implications for interpretable wearable sensing are discussed in Section 5.

\subsection{Prototype learning from smartwatch 3D accelerometer and gyroscope data}

We then evaluated our model on the BasicMotions dataset from the UEA Time Series Repository.
The dataset is 3D accelerometer and 3D gyroscope data collected from a smart watch.
Subjects performed four activities: standing, walking, running, and badminton.
The sampling frequency was 10 Hz and each time series is 10 seconds in length.

\begin{figure*}[t]
    \centering
    \includegraphics[width=\textwidth]{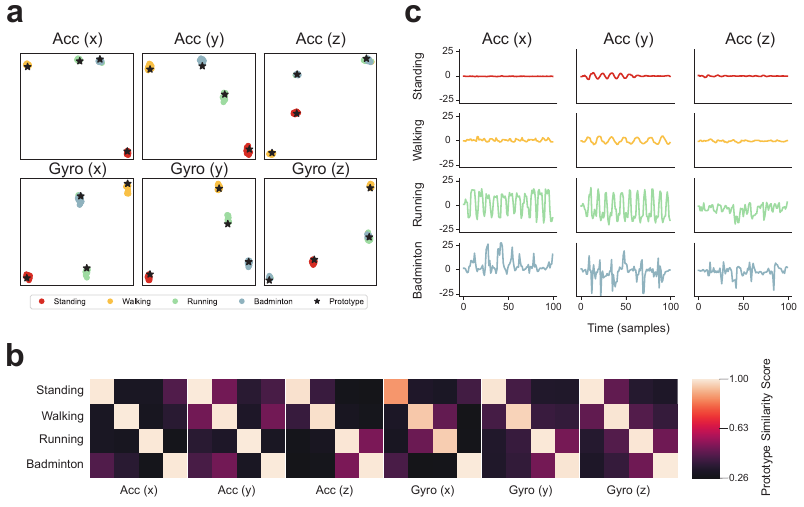}
    \caption{MMPL learns representative examples of all four classes of the BasicMotions dataset. \textbf{a}, UMAP visualization of the learned latent spaces shows the four classes are distinguishable in all variables. \textbf{b}, MMPL selects exactly one univariate prototype per variable when constructing multivariate prototypes. \textbf{c}, Projected prototypes match domain knowledge. The prototype for Running contains periodic signals with high amplitude, while the prototype for Badminton contains high-amplitude signals without periodicity. }
    \label{fig:basicmotions_fig}
\end{figure*}

Accuracy comparisons are shown in Table~\ref{table:uea_accuracies}.
Our model achieves a classification accuracy of 99.8\% on the test set, comparable to state-of-the-art methods.
We observe distinct clusters for each class across all 6 variables as shown in Figure~\ref{fig:basicmotions_fig}a, indicating that all four activities are separable in every sensor channel. This finding is notable because it demonstrates that MMPL's modular design, which processes each variable independently, scales to 6-variable inputs without any degradation in latent space quality or prototype clarity relative to the 3-variable Epilepsy case.
We set the number of univariate prototypes to 4 and initialize them using k-means++.
We observe that each cluster in the latent space has a single univariate prototype assigned to it, confirming that the silhouette-based selection and k-means++ initialization together produce a compact, well-distributed set of prototypes even as the number of variables doubles.
As there are 4 total classes in this dataset, we set the number of multivariate prototypes to 4.
Visualizations of the learned prototypes are presented in Figure~\ref{fig:basicmotions_fig}b.


For all 4 classes, the corresponding multivariate prototype vectors select one univariate prototype per variable, demonstrating that the four activities are fully separable across all six sensor channels simultaneously. The clean one-hot structure of the multivariate prototypes in Figure~\ref{fig:basicmotions_fig}b confirms that the two-stage design produces decomposable, human-verifiable explanations even as the dimensionality of the problem increases.
Projected prototypes are shown in Figure~\ref{fig:basicmotions_fig}c.
The prototype for Running contains periodic signals with high amplitude, while the prototype for Badminton contains high-amplitude signals without periodicity. This distinction, periodicity versus aperiodicity at similar amplitude levels, is a subtle pattern difference that would be difficult to surface with a single shared latent space or a post-hoc saliency method, but is made explicit by MMPL's per-variable prototype structure.



Overall, our results confirm that MMPL scales to 6-variable inputs, maintaining both accuracy and prototype clarity as the number of sensor channels increases. The scalability implications of this finding are discussed in Section~5.

\subsection{Prototype learning for character trajectories}

Next, we investigated the utility of MMPL in prototype learning for character trajectories. The CharacterTrajectories dataset \cite{charactertrajectories} consists of pen tip velocity trajectory data from various character samples written on a tablet.
The dataset only includes 20 characters that have a single "pen down" segment, meaning the subject does not need to lift their pen in order to write the character.
Each sample contains 3 variables: x, y, and pen tip force.
Furthermore, the data are numerically differentiated and Gaussian smoothed.
The classification objective is to predict the character from the trajectory data.

\begin{figure*}[t]
    \centering
    \includegraphics[width=\textwidth]{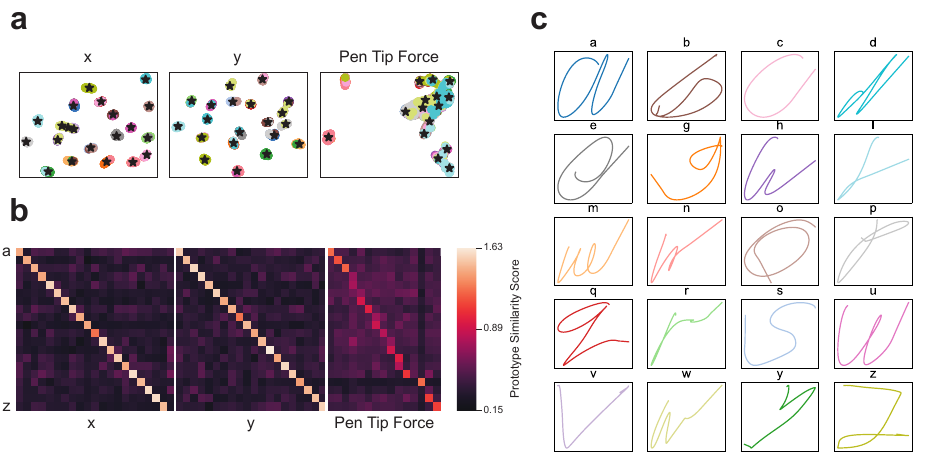}
    \caption{MMPL learns representative examples of all 20 classes of the CharacterTrajectories dataset. \textbf{a}, UMAP visualization of the learned latent spaces shows the 20 classes are distinguishable in the x and y variables, but Pen Tip Force is a poor separator of classes. \textbf{b}, MMPL selects exactly one univariate prototype per variable for the first two variables when constructing multivariate prototypes. Prototypes are selected much more weakly for Pen Tip Force, indicating its relative weakness in terms of distinguishing classes compared to the x and y variables. \textbf{c}, Projected and numerically integrated prototypes match expected appearance of characters.}
    \label{fig:charactertrajectories_fig}
\end{figure*}

We observe a test set accuracy of 96.6\%, while WEASEL+MUSE and MLSTM-FCN achieve accuracies of 99.0\% and 99.3\% respectively as shown in Table~\ref{table:uea_accuracies}
On the CharacterTrajectories dataset, our framework does not sacrifice much accuracy in order to gain interpretable insights on the data.
To verify interpretability, we show UMAP visualizations of the learned latent spaces in Figure~\ref{fig:charactertrajectories_fig}a.
We find that our model is able to separate the 20 characters in the x and y variables, but fails to do so with pen tip force, as multiple classes overlap in the latent space. The lack of cluster structure in the pen tip force latent space is a signal that this variable carries little discriminative information for character identity.
Our model provides insight into feature importance on real-world data: the x and y variables provide important information to distinguish between classes, while pen tip force is less discriminative. This constitutes an automatic feature relevance result in the pattern recognition sense: rather than requiring manual feature engineering or post-hoc attribution analysis, MMPL implicitly identifies which data streams carry discriminative information and which do not, directly from the raw time series \cite{rudin2019stop, samek2021explaining}.

We set the number of univariate prototypes for x, y and pen tip force to 20, 20, and 15 using our silhouette method. The lower prototype count selected for pen tip force, relative to the 20 characters in the dataset, is consistent with the poor cluster structure observed in its latent space, and reflects the silhouette method's ability to adapt the model's expressiveness to the actual information content of each variable.
For variables x and y, our model learns one univariate prototype for each distinct cluster, demonstrating its ability to scale to 20-class problems while maintaining a one-to-one correspondence between prototypes and class-level patterns.
We also set the number of multivariate prototypes to 20 as we are considering 20 characters.
We visualize the multivariate prototypes for CharacterTrajectories in Figure~\ref{fig:charactertrajectories_fig}b.

We observe the "one-hot encoded" pattern for x and y similar to the learned prototypes for the simulated dataset as shown in Figure~\ref{fig:simulated_fig}e.
This means that the multivariate prototypes select exactly one univariate prototype from the x and y variables.
However, the model's selection of univariate prototypes for pen tip force has roughly half the strength of the x and y selections. This graded selection strength means that the multivariate prototypes encode which pattern combination defines a class, as well as how strongly each variable contributes to that definition. This provides a soft, continuous measure of variable importance rather than a binary include/exclude decision.
We also project the learned prototypes into the input space and numerically integrate them in order to visualize the characters, as shown in Figure~\ref{fig:charactertrajectories_fig}c.
Our model correctly identifies a prototype representative of each class label, and the projected characters are visually recognizable, confirming that the nearest-neighbor projection preserves the essential shape information encoded in the latent prototype.



This result highlights one of MMPL's most practically useful properties: automatic, implicit sensor relevance identification at the stream level rather than the time-point level. The broader significance of this finding for the pattern recognition community is discussed in Section~5.

\subsection{Prototype learning from biosensing measurements during affective states}

Finally, we investigated MMPL's ability in prototype learning during affective tasks and from a comprehensive multi-modal biosensing dataset. The Wearable Stress and Affect Detection (WESAD) \cite{wesad} dataset consists of biosensing measurements during affective states including stress, amusement, and a neutral baseline. 
Specifically, the dataset features both physiological and accelerometer data, collected from both the chest and the wrist of participants as they experienced different affective states (e.g., stressed, amusement).
The dataset features two classification tasks: the binary task of detecting stress, and the multi-class task of differentiating between a baseline state, stress, and amusement.

We specifically focus on the multi-class classification task using all available chest data.
This includes the following modalities: electrodermal activity (EDA), electromyogram (EMG), electrocardiogram (ECG), respiration, body temperature, and tri-axial accelerometer data in 3 dimensions.
We pre-process the biosensor signals according to the following procedure. We first de-mean input signals and then apply appropriate low-pass/high-pass filters per-modality according to the procedure specified in the original WESAD paper. We then decimate input signals to a lower frequency of 20 Hz and finally segment input signals with a window of 60 seconds and a stride of 1 second.
Unlike the original paper, we do not rely on manually extracted features per input segment.
Instead, we feed the entire temporal segment into our model and treat the learned embeddings from the encoder stage as the extracted features.

On the task of classifying multiple affective states using all available chest data, we observe a test set accuracy of 78.8\%, which is slightly higher than the highest accuracy reported across multiple standard machine learning methods (76.5\%) in the original dataset paper \cite{wesad}. We note that a direct comparison to deep learning baselines on WESAD is not available in the published literature using this preprocessing protocol; comparison against LSTM-based and transformer-based deep learning baselines is an important direction for future work. As we will show, MMPL can attain high accuracy while also providing a high level of interpretability.

\begin{figure*}
    \centering
    \includegraphics[width=\textwidth]{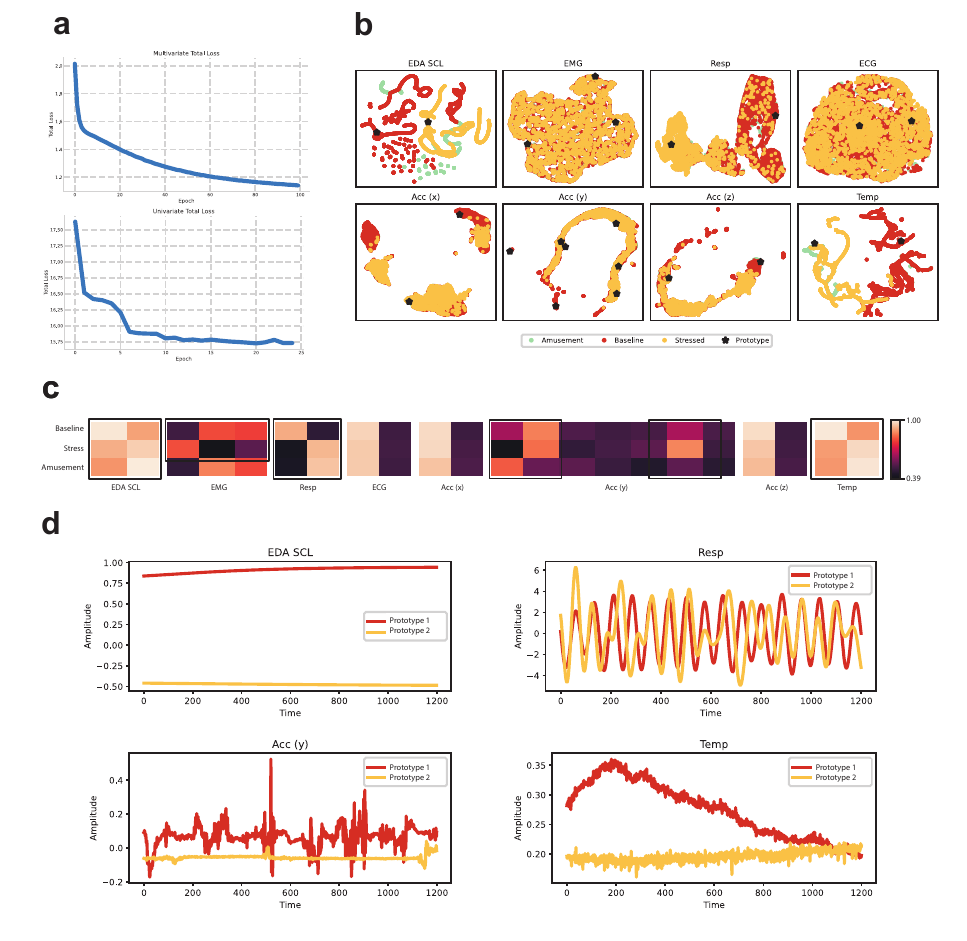}
    \caption{MMPL learns multivariate prototypes for complex biosensing data. \textbf{a}, Loss curves for multivariate and univariate training stages, demonstrating the convergence of our classifier. \textbf{b}, The learned latent representations of the input segments shows that certain modalities are more useful for classification than others: EDA, Temperature, and the accelerometer data are able to separate data into clusters, while EMG and ECG struggle to form any separation. \textbf{c}, Learned multivariate prototypes are much more complex than just one highly selected prototype per class. MMPL finds correlation between Stress and Amusement in the EDA, Respiration, and Temperature modalities, and separates the Baseline class from the others. In the EMG modality, Stress is differentiated while in the Acc (y) modality, Amusement is differentiated. \textbf{d}, Projected prototypes for select modalities are shown. Here, prototype 1 corresponds to stress and prototype 2 corresponds to baseline.} 
    \label{fig:wesad_fig}
\end{figure*}

Both the univariate and multivariate training stages converge smoothly 
(Figure~\ref{fig:wesad_fig}a), confirming that our sequential training 
procedure, freezing the univariate modules before optimizing the 
multivariate prototype layer, is stable.

The UMAP visualizations of the per-modality latent spaces 
(Figure~\ref{fig:wesad_fig}b) reveal a hierarchy of modality 
utility. EDA, Temperature, and the accelerometer channels form 
identifiable cluster structure especially for stress vs. baseline, while EMG and ECG fail to produce well-separated clusters. This is a notable departure from the original WESAD paper \cite{wesad}, which identifies accelerometer data as a poor separator of affective states based on hand-crafted feature 
accuracy. The discrepancy suggests that end-to-end LSTM encoding 
extracts discriminative temporal dynamics from accelerometer signals 
that are not captured by fixed summary statistics. For several modalities, 
MMPL requires more univariate prototypes than the number of classes, 
reflecting within-class variability in physiological signals 
across subjects and recording windows. It is notable that the model was not able to identify a reliable single modality prototype for amusement, suggesting a need for more complex interactions to explain this emotion.

The multivariate prototype heatmap (Figure~\ref{fig:wesad_fig}c) reveals distributed, cross-modality patterns, an evidence for the difficulty of affect discrimination task using single modality. Stress and Amusement 
share prototype selections in the EDA, Respiration, and Temperature 
modalities, suggesting that these channels capture a general 
physiological arousal response common to both states rather than 
features specific to either.

We then investigated projected prototypes across select modalities (Figure~\ref{fig:wesad_fig}d). Here, prototype 1 corresponds to stress and prototype 2 corresponds to baseline. The prototypes across EDA, accelerometer, and temperature show distinct amplitude trajectories between stress and baseline, consistent with known differences in skin conductance, expected movement, and body temperature in high arousal affective state.

Our findings demonstrate that MMPL can simultaneously exceed the 
accuracy of hand-crafted feature pipelines and provide interpretable, 
modality-level prototype representations for complex real-world 
biosensing data. The broader implications of the WESAD modality 
selection findings for affect detection research are discussed in 
Section~5.

\section{Discussion and Conclusion}

Across five datasets spanning a simulated data, physical activity recognition, handwriting trajectory classification, and real-world multi-modal biosensing measurements, MMPL consistently recovers the underlying structure of the data through its learned prototype representations. The ablation study on the simulated dataset quantifies the contribution of each architectural choice: variable separation alone improves the accuracy from 54.2\% to 74.8\% over the single-stage baseline, and adding contrastive pre-training increases this further to 98.5\%. The failure of a single shared latent space to encode heterogeneous multi-variable patterns, as demonstrated by the one-stage model's accuracy, provides evidence that variable separation may be a functional necessity for this class of data. The further performance gap between the two-stage model trained with and without contrastive pre-training suggests that the quality of the latent space is a prerequisite for meaningful prototype learning: without well-separated clusters, prototype redundancy increases and multivariate prototype interpretation becomes ambiguous. This finding aligns with broader trends in representation learning \cite{chen2019looks}, where structured pre-training consistently improves downstream task performance.

On the real-world datasets, MMPL's prototype representations align consistently with domain knowledge. On accelerometer data during seizure mimicking behavior, the four univariate prototypes learned per accelerometer axis correspond to the four activity classes, and the projected multivariate prototypes recover the expected biomechanical distinctions. On the BasicMotions dataset, the same clean one-to-one prototype structure holds across all six accelerometer and gyroscope channels, suggesting scale to higher number of modalities. The CharacterTrajectories results reveal MMPL's ability to perform implicit feature selection at the variable level: the multivariate prototypes select strongly from the x and y trajectory variables but weakly from pen tip force, providing an automatic data-driven assessment of which sensor channels carry discriminative information. This stream-level relevance identification goes beyond standard feature attribution methods \cite{samek2021explaining}. It also has direct implications for sensor selection and data collection strategy in future multi-modal studies. Finally, on the WESAD dataset, MMPL identifies prototypes across various biosensing measurements, suggesting that end-to-end prototype learning can both outperform and explain itself on complex physiological data. This is particularly of interest because multi-modal biosensing domain where black-box models are frequently met with skepticism from clinical stakeholders.

For the pattern recognition community, MMPL occupies a practically important position between fully opaque high-accuracy models and highly interpretable but less accurate feature-based methods. Post-hoc explanation approaches such as SHAP \cite{chen2023algorithms} or GradCAM \cite{selvaraju2017grad} approximate explanations for a fixed black-box model and are not guaranteed to reflect the model's true decision logic. Unlike these approaches, MMPL's interpretability is intrinsic: the prototypes are identified from the model's decision boundaries, and every prediction is made by explicit comparison to learned exemplars. The modular architecture further allows each variable's encoder to be independently replaced with a task-appropriate architecture. This offers a practical solution for the heterogeneous sensor fusion problems common in clinical monitoring, industrial fault detection, and environmental sensing.

Our findings suggest that MMPL struggles with stationary, high-frequency modalities such as EMG and ECG, where class-label-based contrastive pre-training does not produce well-separated latent clusters. This is likely because these signals have similar statistical characteristics across affective states. This is reflected in the low and non-varied prototype similarity scores for the EMG and ECG modalities in Figure \ref{fig:wesad_fig}b. The CharacterTrajectories accuracy of 96.6\% lags behind WEASEL+MUSE (99.0\%) and MLSTM-FCN (99.3\%), suggesting that the LSTM encoder may not be optimal for smooth, continuous trajectory-type data where convolutional or attention-based encoders may be more appropriate \cite{zerveas2021transformer}

Our future work includes replacing the LSTM encoder with temporal convolutional networks or transformer-based encoders \cite{zerveas2021transformer} for high-frequency modalities. These architectures are better suited to capturing the spectral characteristics of signals such as EMG and ECG. The use of class labels to determine contrastive pair similarity is a straightforward but potentially suboptimal choice; future work should investigate whether self-supervised similarity criteria based on temporal neighborhood, augmentation invariance, or nearest-neighbor distance produce better-structured latent spaces. Finally, replacing the deterministic LSTM encoder with a variational autoencoder would enable probabilistic prototype placement and uncertainty quantification in predictions, which is of direct relevance to clinical decision support applications.

For researchers and practitioners in pattern recognition, biosensing, and related fields, MMPL offers a practical path to interpretable multi-sensor classification without sacrificing competitive predictive accuracy. The framework's modular architecture accommodates the signal heterogeneity present in real-world multi-modal sensor systems, allowing different encoder architectures, latent space sizes, and prototype counts per variable. The combination of competitive accuracy, intrinsic interpretability, and modular design positions MMPL as a useful framework for pattern recognition applications in clinical decision support, wearable health monitoring, environmental sensing, and industrial process control, where model transparency is a prerequisite for deployment alongside predictive performance.

\section*{Acknowledgment}
We gratefully acknowledge the feedback and comments provided by the members of the Abbasi Lab and the Neuroscape Center at UCSF.

\section*{Declaration of interests}
The authors declare no competing financial interests.

\section*{Data and Code Availability}
The code for MMPL is publicly available at \url{https://github.com/abbasilab/MMPL}. The simulated dataset used in this study is available in the repository. The three UEA benchmark datasets (Epilepsy, BasicMotions, and CharacterTrajectories) are publicly available through the UEA Multivariate Time Series Classification Archive at \url{https://timeseriesclassification.com}. The WESAD dataset is publicly available and can be accessed at \url{https://archive.ics.uci.edu/dataset/465/wesad+wearable+stress+and+affect+detection}.

\bibliographystyle{unsrt}
\bibliography{references}

\end{document}